\documentclass[letterpaper]{article}
\usepackage{iccc}
\usepackage{graphicx} 
\usepackage{url}
\usepackage{booktabs}
\usepackage{multirow}

\usepackage{times}
\usepackage{helvet}
\usepackage{courier}
\title{Let the Poem Hit the Rhythm: Using a Byte-Based Transformer for Beat-Aligned Poetry Generation}

\author{Mohamad Elzohbi, Richard Zhao\\
Department of Computer Science\\
University of Calgary\\
Calgary, Alberta, Canada, T2N 1N4\\
\{melzohbi, richard.zhao1\}@ucalgary.ca\\
}
\begin{document} 
\maketitle
\begin{abstract}
\begin{quote}

The intersection between poetry and music provides an interesting case for computational creativity, yet remains relatively unexplored. This paper explores the integration of poetry and music through the lens of beat patterns, investigating whether a byte-based language model can generate words that fit specific beat patterns within the context of poetry. Drawing on earlier studies, we developed a method to train a byte-based transformer model, ByT5, to align poems with beat patterns. The results demonstrate a high level of beat alignment while maintaining semantic coherence. Future work will aim to improve the model's ability to create complete beat-aligned poems.

\end{quote}
\end{abstract}

\section{Introduction}

The integration of poetry and music has a significant impact on the aesthetic reception of both art forms. This intersection is evident in the shared terminology, such as “rhythm”, which plays a vital role in cultural transmission and emotional expression. Despite this similarity, limited comparative studies have explored rhythms in music as opposed to rhythms in poetry. This scarcity can mainly be attributed to the complexity of the involvement of experts in both domains \cite{patel2010music}. The patterns and sounds of words are generally overlooked when studying poetry, whereas studies tend to concentrate on other language-related aspects \cite{hirjee2009automatic}.

Songs are great examples of creative language meeting music. They often stick to lyrics that fit well with the musical tones and beats forming patterns that not only convey figurative meaning but also align smoothly with the music. Linguistic stress is comparable to musical accents, indicating points of emphasis. \citeauthor{lerdahl2001sounds} (\citeyear{lerdahl2001sounds}) explored the relationship between the sounds of English poetry and music, focusing on how stress patterns and pitch levels in poetry and music are related. In a later study, \citeauthor{lerdahl2013musical} (\citeyear{lerdahl2013musical}) used the Beatles' song ``Yesterday" as a key example to demonstrate these connections. The challenge in generating lyrics is in balancing coherence with fitting into a steady musical rhythm. While many studies focus on rhyme, they often overlook the essential rhythmic component of the lyrics \cite{xue2021deeprapper}. However, some research has explored generating lyrics of various languages that align with music from stress and pitch perspective \cite{oliveira2007tra,sheng2021songmass,chen2024scansion}.

Additionally, beat rhythm is another point of comparison between language and music, though whether the language itself is rhythmic has been a subject of debate and research has led to controversial findings \cite{rathcke2021tapping}. The African talking drums provide a fascinating case study of the convergence of sound and language. These drums are used to mimic the tonal and rhythmic patterns of spoken language, acting as acoustic speech surrogates illustrating the relationship between rhythmic patterns and verbal communication \cite{ong1977african}. Similarly, Arabic poetry has deep ties to drum music, serving as the foundation for Arabic prosody. It is said that the rhythmic pounding of hammers in the street of braziers in al-Baṣrah inspired al-Khalīl al-Farāhīdī to establish the rules of Arabic prosody based on the alignment he found between the clanging patterns and the speech rhythm in poetry \cite{ibnMutaz}. 

\citeauthor{allen1972} (\citeyear{allen1972}) studied the perceptual beat locations in words taken from spontaneous English conversations. The study involved several experiments including having participants tap their fingers along with perceived beats in the words, confirming that rhythmic beats often align with vowel onsets, i.e. the transitions from consonants to vowels, and were more accurately detected in stressed syllables. \citeauthor{rathcke2021tapping} (\citeyear{rathcke2021tapping}) recently revisited the topic and studied the validity of sensorimotor synchronization to perceive speech rhythms where similar tapping experiments showed that people tend to identify perceptual beats at vowel onsets in English sentences. These studies mainly focus on English, but a similar concept can be observed in other languages such as Arabic, where scansion places rhythmic beats on vocalized consonants (those followed by short vowels), which characterize the meter of the poem \cite{frolov2021classical}. 

\begin{figure*}[!htb]
	\centering
	\includegraphics[width=0.6\textwidth]{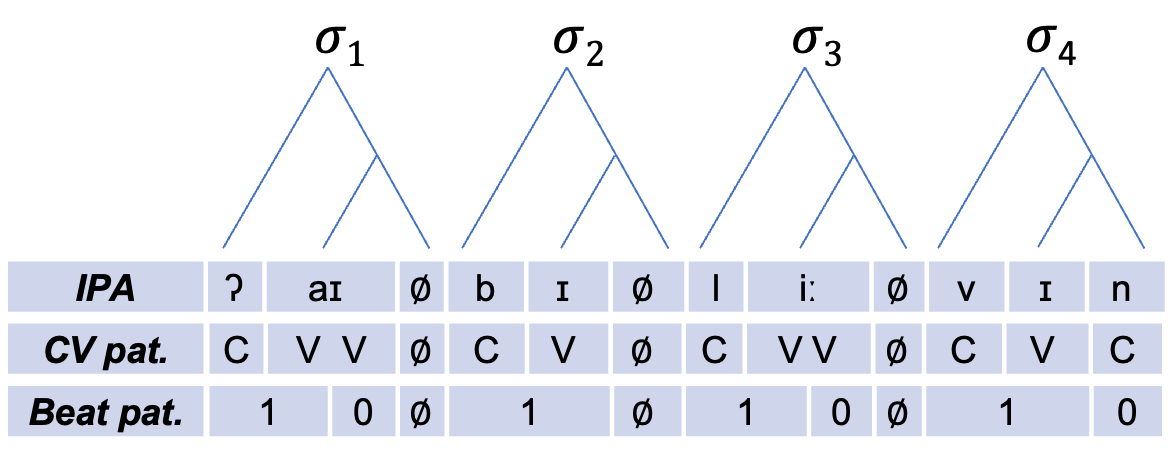}
        \caption{The process of converting a four-syllable phrase \textit{``I believe in"} from its phones to a beat pattern.}
	\label{fig:IPA}
   	% \vspace{-9pt}
\end{figure*}

This raises several research questions: Can a large language model effectively learn beat patterns perceived from natural language? How accurately can such a model generate textual content that conforms to predetermined rhythmic constraints? And, importantly, does this capability extend to maintaining semantic coherence while adhering to these constraints? 

Building on these questions, this short paper seeks to determine if a large language model can be trained to select words that align with a given beat pattern. Our focus is on generating words in the context of poems, aiming to make them resemble lyrics used in music. To achieve this, we develop a model that maps graphemes to beat patterns, building on insights from earlier research. We then utilize a byte-based transformer model to replace or insert words that align with a specified beat pattern, all without exposing the model to phonemes or audio-based data, but instead by interpreting symbolic beat patterns derived from the arrangement of consonants and vowels. To measure the model's quality, we employ automated evaluation metrics. 

\section{Methodology}

In this study, we selected the ByT5 model \cite{xue2022byt5}, a byte-level transformer model derived from the T5 architecture which is an encoder-decoder transformer developed by Google \cite{raffel2020exploring}. Unlike models that break text into subword-level units, ByT5 operates at the character level, providing the precision needed to handle text with granular control over character-level patterns. 

\subsection{Task Formalization} 
The task involves inserting a set of words $W' = (w'_1, w'_2, \ldots, w'_i)$ within a given poetry verse $S = (w_1, w_2, \ldots, w_n)$, $i < n$, that exhibit a given beat, $B(W')$. The task may also be formalized as replacing a set of words $W \subset S$ with an undesirable beat $B(W)$ with a set of words $W'$ of desired beat $B(W')$. This task requires a poetry dataset as well as a graphemes-to-beat transformation $B(.)$. 

\subsection{Dataset} 
The dataset used in this study is derived from the English-labeled subset of a poetry corpus compiled by \citeauthor{haider-2021-metrical} (\citeyear{haider-2021-metrical}). To ensure the dataset contained only English poems, we applied additional cleaning using an XLM-RoBERTa-based language identification model trained on a language identification dataset\footnote{The model is available at: \url{https://huggingface.co/papluca}}. This step filtered out non-English poetry verses, leaving only the relevant poetry verses for our analysis. We used only the poetry verses from the dataset ignoring non-relevant information. 

\begin{figure}
	\centering
	\includegraphics[width=0.35\textwidth]{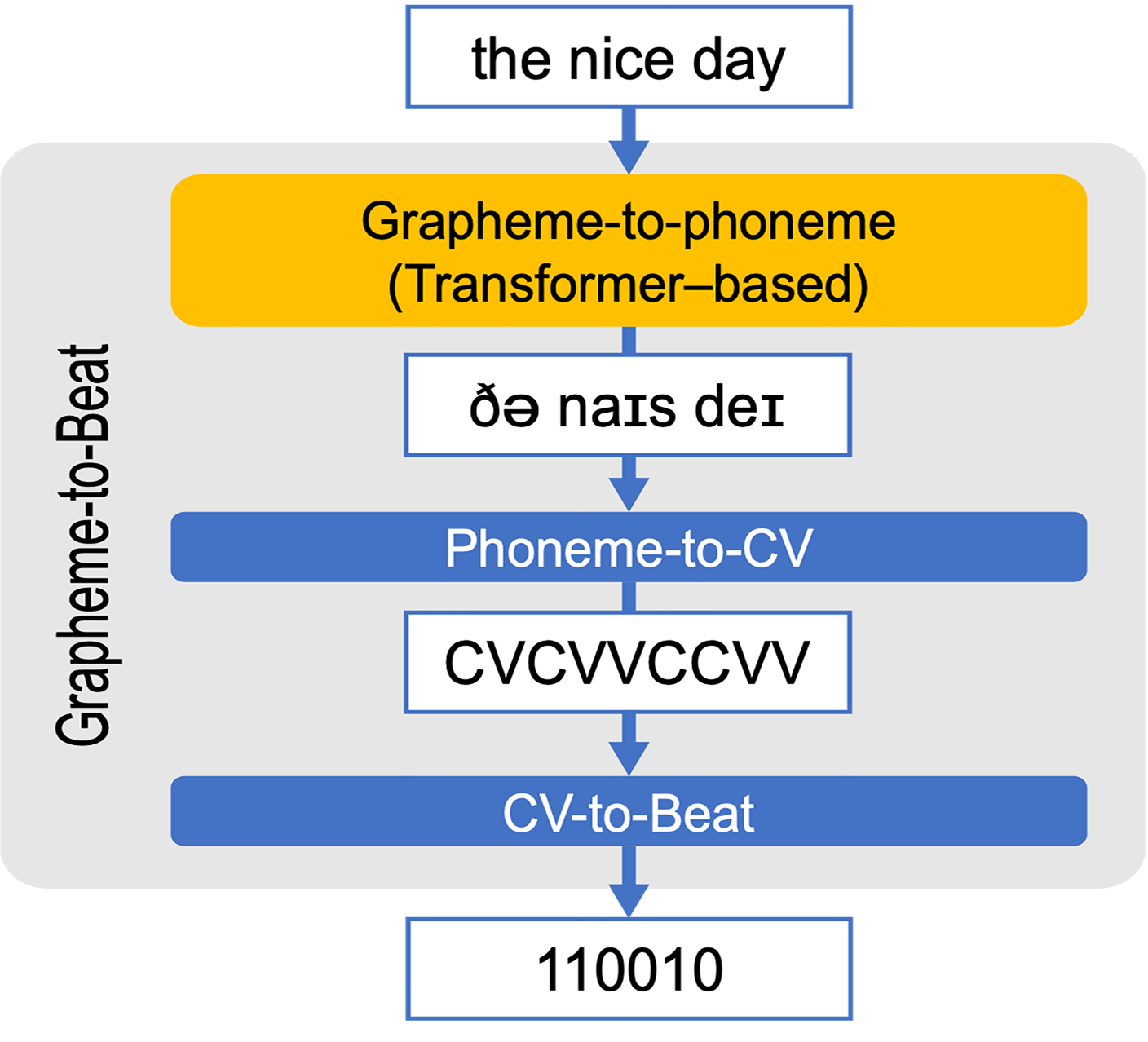}
        \caption{Grapheme-to-Beat pattern model.}
	\label{fig:CV}
   	% \vspace{-9pt}
\end{figure}

\begin{figure*}[!htb]
	\centering
	\includegraphics[width=\textwidth]{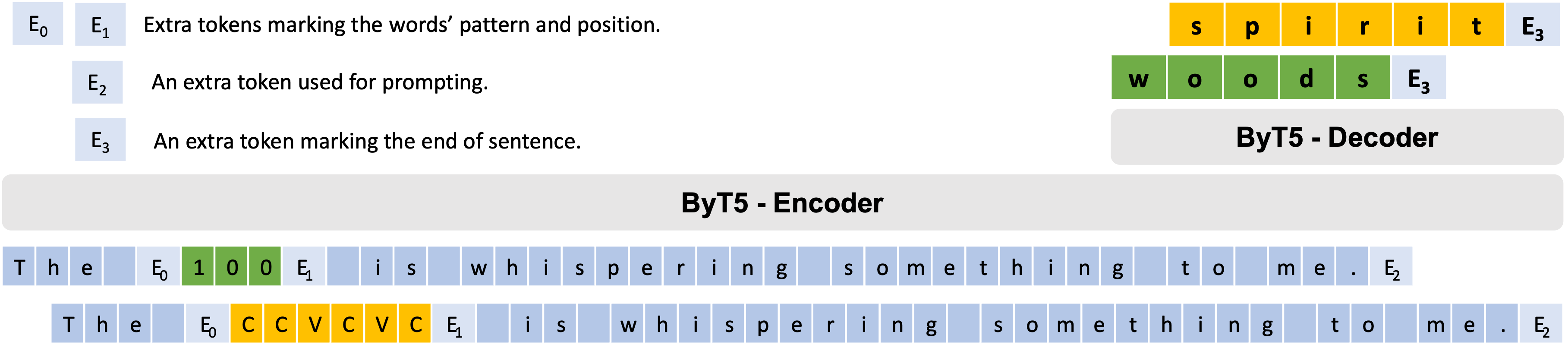}
        \caption{An example input/output of the ByT5-B in green and ByT5-CV in orange.\nocite{*}}
	\label{fig:CV-pattern}
   	% \vspace{-9pt}
\end{figure*}

\subsection{Grapheme-to-Beat Transformation} 
To identify the rhythmic beats in English words, we follow the findings outlined in past studies \cite{allen1972,rathcke2021tapping,Hermes2023}. That is, we place the beat attack on the transition from a consonant to a vowel. A syllable comprises an onset, a nucleus, and a coda. The nucleus is typically a vowel- either a short vowel, long vowel, or diphthong- while the onset and coda can include zero or more consonants. It is possible for a syllable in English to have a null onset or coda, but if there is no onset, a glottal stop generally precedes the vowel \cite{Hermes2023}, which can still be considered a vowel onset. Any other sounds — such as consonants not followed by a vowel, vowel elongations, or the second vowel in a diphthong — are treated as non-beat units (or rests) (Figure \ref{fig:IPA}).

English script is orthographic deep, where there is little one-to-one correspondence between letters and sounds. To get the consonant-vowel pattern $CV(w)$ for a given word $w$, we first need to extract its phonemes from the English script. To do this, we used a pre-trained grapheme-to-phoneme transformer model\footnote{\url{https://github.com/as-ideas/DeepPhonemizer}}. This model transforms text into phonemes, which we then convert to consonants and vowels with one-to-one mapping. Long vowels and diphthongs are mapped to double vowels ``VV", while consonants are denoted by ``C". We then convert these CV patterns into beat patterns: a ``1" represents a beat where there is a vowel onset, and a ``0" represents a non-beat unit (or rest) (Figure \ref{fig:CV}).

\subsection{Model Training} 

We fine-tuned the pre-trained ByT5 model on the processed dataset to generate words that align with a specified rhythmic beat pattern. During training, we used a masking strategy to simulate the task's objective. For each verse, let us say $S = (l_1, l_2, ..., l_n)$ where $l$ represents an individual letter in the verse, we randomly masked a set of words $W = (l_i, l_{i+1}, ..., l_{i+j-1})$ where $i$ is the position of the first letter and $j$ is the total letter count in $W$. 

To indicate where the masking occurred in $S$, we used special tokens ($E_0$ and $E_1$) so that the masked version of $S$ looks like this: $S' = (l_1, ..., E_0, B(W), E_1, ..., l_n)$. $B(W)$ represents the beat pattern of the original words. We then appended a special token $E_2$ followed by $W$ to prompt the model to align the words with their corresponding beats (Figure \ref{fig:CV-pattern}). For a comparative analysis, we used a similar strategy to fine-tune a model that focused on consonant-vowel patterns with $CV(W)$ replacing $B(W)$ in $S'$.
 
To ensure the model could handle spans of varying lengths, we applied a span-masking strategy inspired by SpanBERT \cite{joshi2020spanbert}. This strategy involved sampling span lengths from a geometric distribution with a probability parameter of $0.2$.  A maximum threshold was set at $25\%$ of verse words' count, but the geometric distribution favours shorter spans.

\section{Experimental Setup}

\subsection{Dataset Split} 
Our dataset consists of $1,038,743$ verses obtained after data preprocessing. We excluded potentially garbled lines, as was done in the \textsc{Quatrain} dataset \cite{belouadi2023bygpt5}. We then split the dataset into two random subsets: a training set comprising $1,033,549$ examples and an evaluation set with $5,194$ examples. 

% Quatrain dataset has $671903$.
\subsection{Training Setup} 
We fine-tuned two ByT5-base models on the specified tasks: \textbf{ByT5-B} was trained with a focus on beat patterns, while \textbf{ByT5-CV} focused on consonant-vowel patterns. We also fine-tuned two ByT5-base models \textbf{ByT5-CV-base} and \textbf{ByT5-B-base} where attention masks were set to zero in the pattern positions, ensuring that the model did not attend to these patterns, using this configuration as a baseline. All models were trained for four epochs on an NVIDIA A100 GPU. The learning rate was set to $3e-4$ with a cosine scheduler and a weight decay of $0.1$. The training batch size was set to $128$, and the evaluation batch size was set to $16$.

\subsection{\textsc{GPT-4} Baseline}
To compare our results against the current state-of-the-art large language models, we used \textsc{GPT-4} (version: \texttt{gpt-4-0613}), which is OpenAI's most capable model at the time of this research. We randomly selected $1000$ examples from our dataset and masked them as described. We prompted \textsc{GPT-4} with a short explanation of the problem and a few-shot examples to evaluate its performance on reconstructing beat patterns\footnote{The source code, the dataset, and \textsc{GPT-4} prompts used are available at: \url{https://github.com/melzohbi/poem-rhythm}}. At the time of this research, we did not have access to any fine-tuning capabilities of \textsc{GPT-4}. However, the purpose of this comparison was to assess whether \textsc{GPT-4}, in its current state, could understand and perform this task effectively, given its proficiency in various downstream tasks. We acknowledge that fine-tuning is important for a fairer comparison, and we plan to compare our model with a fine-tuned version of \textsc{GPT-4} in the future if such capabilities become available.

\subsection{Automated Evaluation Metrics}
The task's primary objective is to generate words that match a beat pattern while maintaining semantic coherence within the context of a poetry verse. Using automated evaluation, we measure Coherence and Beat Alignment (while human-based evaluations will be sought in future work).

\paragraph{Coherence Score} measures how well the predicted words fit with the surrounding context. To evaluate coherence, we utilized the original subword-level T5 model (using the base-size version) to derive a log-perplexity score (through calculating the cross-entropy loss). We inserted a special token at the location of the masked span in the verse, so the sequence became \(S = (w_1, \ldots, E_0, \ldots, w_n)\). We then prompted the model to predict the tokens replacing the special $E_0$ token, computing the cross-entropy loss for the prediction. This pre-trained model, serving as an external scoring mechanism, was independent of our dataset. This setup allows us to verify whether or not our training process impacts the coherence of the model.

\paragraph{Alignment Scores} evaluate the alignment of generated words with the required beat rhythm. We employed two metrics:
\begin{itemize}
    \item \textit{Exact Alignment Accuracy:} This metric checks whether the generated word precisely aligns with the expected beat rhythm, resulting in a binary outcome (0 for non-alignment, 1 for exact alignment).
    \item \textit{Levenshtein distance:} This metric assesses alignment by calculating the normalized Levenshtein distance between the generated word and the expected beat rhythm. This measure accounts for cases where the alignment may not be exact but is reasonably close, providing additional flexibility in evaluation. 
\end{itemize}

\section{Results}

This section explores how well the fine-tuned ByT5 models meet the task's objectives. We evaluate their performance in terms of Coherence and Beat Alignment. In the following section, we discuss the implications of these results for further research.

\paragraph{Coherence:} The results in Table \ref{table:1} show that both models achieved a comparable log perplexity with the baseline model, with the ByT5-CV version slightly ahead with a log-perplexity score of $7.48$, compared to $7.494$ for the ByT5-B version. The baseline models score between $7.442$ and $7.432$. These outcomes suggest that the custom fine-tuning process didn't significantly impact the model's semantic coherence, indicating that it adapted well to the rhythmic task without sacrificing fluency. We did not evaluate GPT-4 output for coherence, given that T5 exhibits lower fluency compared to GPT-4.

\paragraph{Alignment Scores:} 
Both of our models showed a high level of beat alignment accuracy. The ByT5-CV model achieved an Exact Alignment Accuracy score of $98.88\%$, while the ByT5-B variant achieved $98.31\%$. The Levenshtein distance, which allows for some flexibility, indicated near-perfect scores of $99.83\%$ and $99.63\%$ respectively, suggesting that the models generally aligned with the rhythm, with only minor deviations. This demonstrates that the models understand rhythmic patterns even when they do not exactly match. In contrast, the baseline ByT5 models, including \textsc{GPT-4} on the $1000$ examples, had much lower scores for both exact alignment and Levenshtein distance, indicating that these patterns are challenging to learn without proper training.

These findings confirm that our fine-tuned ByT5 models can generate words that align with beat patterns while maintaining coherence. This sets the stage for further research into generating entire verses or poems with consistent rhythmic structures.

\begin{table}[!htb]
    \centering
    \begin{tabular}{l|c|c|c}
    \toprule 
        Model & Coherence & Accuracy & Levenshtein   \\ \midrule
        ByT5-Base-CV & 7.442 & 0.3512 & 0.7670 \\ 
        ByT5-Base-B & \textbf{7.432} &  0.5575 & 0.8323 \\ 
        \textsc{GPT-4} & - & 0.2550 & 0.6105 \\ \midrule
        ByT5-CV & 7.480 & \textbf{0.9888} & \textbf{0.9983} \\ 
        ByT5-B & 7.494 & 0.9831 & 0.9963  \\ \bottomrule
    \end{tabular}
    \caption{Performance comparison between our consonant-vowel model, beat model and the baselines. For Coherence, lower is better. For Exact Alignment Accuracy and Levenshtein distance, higher is better.}
\label{table:1}
\end{table}

\section{Conclusion and Future Work}
In this study, we investigated the capabilities of ByT5, a byte-level language model, particularly its aptitude for generating words that conform to specific consonant-vowel patterns and rhythmic beats. Our methodology focused on fine-tuning multiple ByT5-based models on a conditional mask-predict objective to reconstruct words with predetermined consonant-vowel patterns and rhythmic beats. The results demonstrated that our models were able to generate words that align with specified patterns while maintaining semantic coherence. Our models showed high rhythmic alignment accuracy indicating their effectiveness in this task without adversely sacrificing the models' fluency. 

The models have potential applications in various co-creative frameworks, including songwriting, rap lyric generation, and in the process of rhythmic poetry creation. In future work, we aim to expand the model's capabilities to not only insert individual phrases but also generate complete verses or poems that maintain rhythmic coherence throughout. We will also explore the impact of unstressed syllables on beat strength and the precise locations and durations of beats backed by human evaluation. By advancing these aspects, we intend to enhance the model's utility furthering the integration between poetry and music.

\section{Acknowledgements}

We thank the anonymous reviewers for their valuable feedback and members of the Serious Games Research Group at the University of Calgary for their support. We appreciate the scholarship provided by Elsevier’s Artificial Intelligence Journal through the ICCC program committee.

\bibliographystyle{iccc}
\bibliography{iccc}

\begin{thebibliography}{}

\bibitem[\protect\citeauthoryear{Allen}{1972}]{allen1972}
Allen, G.~D.
\newblock 1972.
\newblock The location of rhythmic stress beats in english: an experimental study i.
\newblock {\em Language and Speech} 15(1):72--100.
\newblock PMID: 5073939.

\bibitem[\protect\citeauthoryear{Belouadi and Eger}{2023}]{belouadi2023bygpt5}
Belouadi, J., and Eger, S.
\newblock 2023.
\newblock Bygpt5: End-to-end style-conditioned poetry generation with token-free language models.
\newblock In {\em Proceedings of the 61st Annual Meeting of the Association for Computational Linguistics (Volume 1: Long Papers)},  7364--7381.

\bibitem[\protect\citeauthoryear{Chen and Teufel}{2024}]{chen2024scansion}
Chen, Y., and Teufel, S.
\newblock 2024.
\newblock Scansion-based lyrics generation.
\newblock In {\em Proceedings of the 2024 Joint International Conference on Computational Linguistics, Language Resources and Evaluation (LREC-COLING 2024)},  14370--14381.

\bibitem[\protect\citeauthoryear{Conneau \bgroup et al.\egroup }{2019}]{conneau2019unsupervised}
Conneau, A.; Khandelwal, K.; Goyal, N.; Chaudhary, V.; Wenzek, G.; Guzm{\'a}n, F.; Grave, E.; Ott, M.; Zettlemoyer, L.; and Stoyanov, V.
\newblock 2019.
\newblock Unsupervised cross-lingual representation learning at scale.
\newblock {\em arXiv preprint arXiv:1911.02116}.

\bibitem[\protect\citeauthoryear{Durojaye \bgroup et al.\egroup }{2021}]{durojaye2021music}
Durojaye, C.; Knowles, K.~L.; Patten, K.~J.; Garcia, M.~J.; and McBeath, M.~K.
\newblock 2021.
\newblock When music speaks: An acoustic study of the speech surrogacy of the nigerian d{\`u}nd{\`u}n talking drum.
\newblock {\em Frontiers in Communication} 6:652690.

\bibitem[\protect\citeauthoryear{Frolov}{2000}]{frolov2021classical}
Frolov, D.
\newblock 2000.
\newblock {\em Classical Arabic Verse: History and Theory of `arūd}, volume~21.
\newblock Brill.

\bibitem[\protect\citeauthoryear{Haider}{2021}]{haider-2021-metrical}
Haider, T.
\newblock 2021.
\newblock Metrical tagging in the wild: Building and annotating poetry corpora with rhythmic features.
\newblock In {\em Proceedings of the 16th Conference of the European Chapter of the Association for Computational Linguistics: Main Volume},  3715--3725.
\newblock Online: Association for Computational Linguistics.

\bibitem[\protect\citeauthoryear{Hermes}{2023}]{Hermes2023}
Hermes, D.~J.
\newblock 2023.
\newblock Beat detection.
\newblock In {\em The Perceptual Structure of Sound}. Cham, Switzerland: Springer International Publishing.
\newblock  225--259.

\bibitem[\protect\citeauthoryear{Hirjee and Brown}{2009}]{hirjee2009automatic}
Hirjee, H., and Brown, D.~G.
\newblock 2009.
\newblock Automatic detection of internal and imperfect rhymes in rap lyrics.
\newblock In {\em Proceedings of the 10th International Society for Music Information Retrieval Conference (ISMIR)},  711--716.

\bibitem[\protect\citeauthoryear{Ibn~al Mu'tazz}{1976}]{ibnMutaz}
Ibn~al Mu'tazz, A.
\newblock 1976.
\newblock {\em Tabaqāt al-Shu`arā' al-Muḥdathīn}.
\newblock Transcription, Dar Al-Ma`arif.

\bibitem[\protect\citeauthoryear{Joshi \bgroup et al.\egroup }{2020}]{joshi2020spanbert}
Joshi, M.; Chen, D.; Liu, Y.; Weld, D.~S.; Zettlemoyer, L.; and Levy, O.
\newblock 2020.
\newblock Spanbert: Improving pre-training by representing and predicting spans.
\newblock {\em Transactions of the association for computational linguistics} 8:64--77.

\bibitem[\protect\citeauthoryear{Lerdahl}{2001}]{lerdahl2001sounds}
Lerdahl, F.
\newblock 2001.
\newblock The sounds of poetry viewed as music.
\newblock {\em Annals of the New York Academy of Sciences} 930(1):337--354.

\bibitem[\protect\citeauthoryear{Lerdahl}{2013}]{lerdahl2013musical}
Lerdahl, F.
\newblock 2013.
\newblock Musical syntax and its relation to linguistic syntax.
\newblock {\em Language, Music and the Brain, ed M. A. Arbib}  257--272.

\bibitem[\protect\citeauthoryear{Mesaros and Virtanen}{2008}]{mesaros2008automatic}
Mesaros, A., and Virtanen, T.
\newblock 2008.
\newblock Automatic alignment of music audio and lyrics.
\newblock In {\em Proceedings of the 11th Int. Conference on Digital Audio Effects (DAFx-08)}.

\bibitem[\protect\citeauthoryear{Oliveira, Cardoso, and Pereira}{2007}]{oliveira2007tra}
Oliveira, H. R.~G.; Cardoso, F.~A.; and Pereira, F.~C.
\newblock 2007.
\newblock Tra-la-lyrics: An approach to generate text based on rhythm.
\newblock In {\em Proceedings of the 4th. International Joint Workshop on Computational Creativity}.
\newblock A. Cardoso and G. Wiggins.

\bibitem[\protect\citeauthoryear{Ong}{1977}]{ong1977african}
Ong, W.~J.
\newblock 1977.
\newblock African talking drums and oral noetics.
\newblock {\em New Literary History} 8(3):411--429.

\bibitem[\protect\citeauthoryear{Patel}{2010}]{patel2010music}
Patel, A.~D.
\newblock 2010.
\newblock {\em Music, language, and the Brain}.
\newblock Oxford University Press.

\bibitem[\protect\citeauthoryear{Raffel \bgroup et al.\egroup }{2020}]{raffel2020exploring}
Raffel, C.; Shazeer, N.; Roberts, A.; Lee, K.; Narang, S.; Matena, M.; Zhou, Y.; Li, W.; and Liu, P.~J.
\newblock 2020.
\newblock Exploring the limits of transfer learning with a unified text-to-text transformer.
\newblock {\em Journal of machine learning research} 21(140):1--67.

\bibitem[\protect\citeauthoryear{Rathcke \bgroup et al.\egroup }{2021}]{rathcke2021tapping}
Rathcke, T.; Lin, C.-y.; Falk, S.; and Dalla~Bella, S.
\newblock 2021.
\newblock Tapping into linguistic rhythm.
\newblock {\em Laboratory Phonology} 12(1):1--32.

\bibitem[\protect\citeauthoryear{Sheng \bgroup et al.\egroup }{2021}]{sheng2021songmass}
Sheng, Z.; Song, K.; Tan, X.; Ren, Y.; Ye, W.; Zhang, S.; and Qin, T.
\newblock 2021.
\newblock Songmass: Automatic song writing with pre-training and alignment constraint.
\newblock In {\em Proceedings of the AAAI Conference on Artificial Intelligence},  13798--13805.

\bibitem[\protect\citeauthoryear{Xue \bgroup et al.\egroup }{2021}]{xue2021deeprapper}
Xue, L.; Song, K.; Wu, D.; Tan, X.; Zhang, N.~L.; Qin, T.; Zhang, W.-Q.; and Liu, T.-Y.
\newblock 2021.
\newblock Deeprapper: Neural rap generation with rhyme and rhythm modeling.
\newblock In {\em Proceedings of the 59th Annual Meeting of the Association for Computational Linguistics and the 11th International Joint Conference on Natural Language Processing (ACL-IJCNLP)},  69--81.

\bibitem[\protect\citeauthoryear{Xue \bgroup et al.\egroup }{2022}]{xue2022byt5}
Xue, L.; Barua, A.; Constant, N.; Al-Rfou, R.; Narang, S.; Kale, M.; Roberts, A.; and Raffel, C.
\newblock 2022.
\newblock Byt5: Towards a token-free future with pre-trained byte-to-byte models.
\newblock {\em Transactions of the Association for Computational Linguistics} 10:291--306.

\end{thebibliography}

\end{document}